\documentclass[10pt,twocolumn,letterpaper]{article}

\usepackage[final]{cvpr}      %
\usepackage[utf8]{inputenc} %
\usepackage[T1]{fontenc}    %
\usepackage[pagebackref,breaklinks,colorlinks]{hyperref}
\usepackage{url}            %
\usepackage{booktabs}       %
\usepackage{floatrow}
\usepackage{amsfonts}       %
\usepackage{amsmath}
\usepackage{nicefrac}       %
\usepackage{microtype}      %
\usepackage[dvipsnames,table,xcdraw]{xcolor}         %
\usepackage{graphicx}  %
\usepackage{paralist}
\usepackage{makecell}
\usepackage{adjustbox}
\usepackage{multirow}
\usepackage{colortbl}
\usepackage{pifont}
\usepackage{xspace}
\usepackage{rotating}
\usepackage{wrapfig}
\usepackage{bbm}
\usepackage[accsupp]{axessibility}
\usepackage{subcaption}
\usepackage[font=small]{caption}

\newcommand{\veryshortarrow}[1][3pt]{\mathrel{%
   \hbox{\rule[\dimexpr\fontdimen22\textfont2-.2pt\relax]{#1}{.4pt}}%
   \mkern-4mu\hbox{\usefont{U}{lasy}{m}{n}\symbol{41}}}}

\newfloatcommand{capbtabbox}{table}
\newfloatcommand{capbfigbox}{figure}

\definecolor{Gray}{gray}{0.9}

\newcolumntype{R}[2]{%
    >{\adjustbox{angle=#1,lap=\width-(#2)}\bgroup}%
    l%
    <{\egroup}%
}

\newcommand{\source}{\mathcal{S}}  %
\newcommand{\target}{\mathcal{T}}  %

\newcommand{\numclasses}{K}  %
\newcommand{\model}{h}  %
\newcommand{\modelparams}{\Theta}  %

\def\cvprPaperID{20} %
\def\confName{CVPR}
\def\confYear{2022}
\title{Can domain adaptation make object recognition work for everyone?}

\author{
    \textbf{Viraj Prabhu}\thanks{Work done partially as intern at Salesforce Research.}$^{\,\,,1}$ \qquad
    \textbf{Ramprasaath R. Selvaraju}\thanks{Work done at Salesforce Research.}$^{\,\,,2}$ \qquad
    \textbf{Judy Hoffman}$^1$ \qquad 
    \textbf{Nikhil Naik}$^{3}$ \qquad 
    \\
    $^1$Georgia Tech \qquad
    $^2$Artera AI \qquad
    $^3$Salesforce Research \\
    {\small\texttt{\{virajp,judy\}@gatech.edu} \qquad \texttt{ ram@artera.ai} \qquad \texttt{nnaik@salesforce.com} \qquad 
    }
}

\begin{document}

\maketitle

\vspace{-10pt}
\begin{abstract}
  Despite the rapid progress in deep visual recognition, modern computer vision datasets significantly overrepresent the developed world and models trained on such datasets underperform on images from unseen geographies. We investigate the effectiveness of unsupervised domain adaptation (UDA) of such models across geographies at closing this performance gap. To do so, we first curate two shifts from existing datasets to study the Geographical DA problem, and discover new challenges beyond data distribution shift: context shift, wherein object surroundings may change significantly across geographies, and subpopulation shift, wherein the intra-category distributions may shift. We demonstrate the inefficacy of standard DA methods at Geographical DA, highlighting the need for specialized geographical adaptation solutions to address the challenge of making object recognition work for everyone.
\end{abstract}
\vspace{-10pt}

\section{Introduction}
\label{sec:intro}

    As deep-learning based computer vision systems gain widespread adoption, it is crucial that they perform equitably across diverse geographical deployments. However, prior work~\cite{shankar2017no} has found that in practice modern computer vision datasets significantly overrepresent the developed world and models trained on such datasets systematically underperform on images from the rest of the world~\cite{de2019does} (see Fig.~\ref{fig:teaser}). Labeling images from target geographies is a natural solution but may be expensive and difficult to scale. Unsupervised domain adaptation~\cite{saenko2010adapting,ganin2015unsupervised,prabhu2021sentry} (UDA) has extensively studied the problem of adapting models trained on a labeled source to an unlabeled target domain. However, UDA typically considers specific kinds of shifts in data generating distributions (\emph{e.g.} synthetic to real data~\cite{peng2018visda}, or clipart to sketch images~\cite{peng2019moment}), rather than distribution shifts across space and time in the real world.
    In this work, we investigate the effectiveness of UDA techniques at the practical application of adapting trained object recognition models to novel geographies.
    
    Geographical domain adaptation presents two novel challenges beyond shifting data distributions: \emph{context shift} and \emph{subpopulation shift}. Context shift arises from a change in visual context for a given category across geographies (\emph{e.g.} predominantly indoor v/s outdoor basketball courts). Subpopulation shift arises from a change in within-category data distributions (\emph{e.g.} for a `toothbrush' category, the relative proportion of electric v/s mechanical varieties may change across geographies). In our experiments, we demonstrate the inefficacy of conventional adaptation strategies in addressing these additional challenges.
    
    Some prior work has studied the problem of transferring deep visual models to new geographies. De Vries~\emph{et al.}~\cite{de2019does} benchmark the drop in performance of publicly available vision API's on images from diverse geographies from the Dollar Street dataset~\cite{dstreet}, but do not propose a mitigation strategy. Recently, Dubey~\emph{et al.}~\cite{dubey2021adaptive} formulate this as a domain generalization problem and propose a solution that makes use of auxiliary target domain embeddings. We instead pose the problem as one of \emph{domain adaptation} so as to leverage the full potential of  unlabeled target data by allowing model updates on it. We make the following contributions:
    
    \begin{figure}[t]
        \centering
        \includegraphics[width=1.0\linewidth]{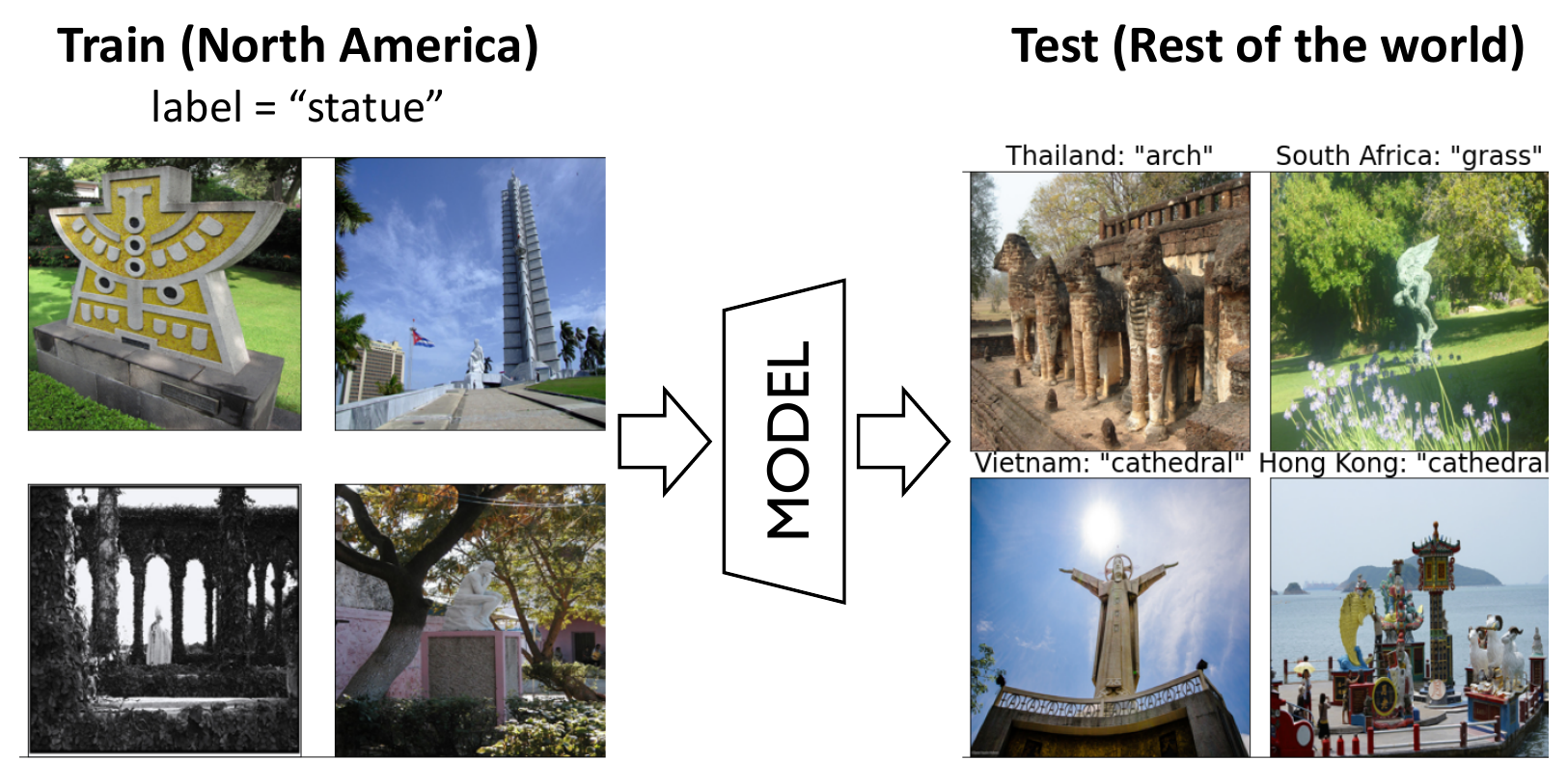}     
        \vspace{-5pt}   
        \caption{Modern computer vision datasets overrepresent the developed world~\cite{shankar2017no}. This leads object recognition models trained on them (left) to underperform on images from novel geographies~\cite{de2019does} (right -- we show country of origin and model prediction above each image). In this work we investigate the effectiveness of domain adaptation~\cite{saenko2010adapting} methods in bridging this performance gap.}
        \label{fig:teaser}
      \end{figure}

      \begin{figure*}[t]
        \centering
        \begin{subfigure}[t]{0.9\textwidth}
            \vskip 0pt 
          \includegraphics[width=\textwidth]{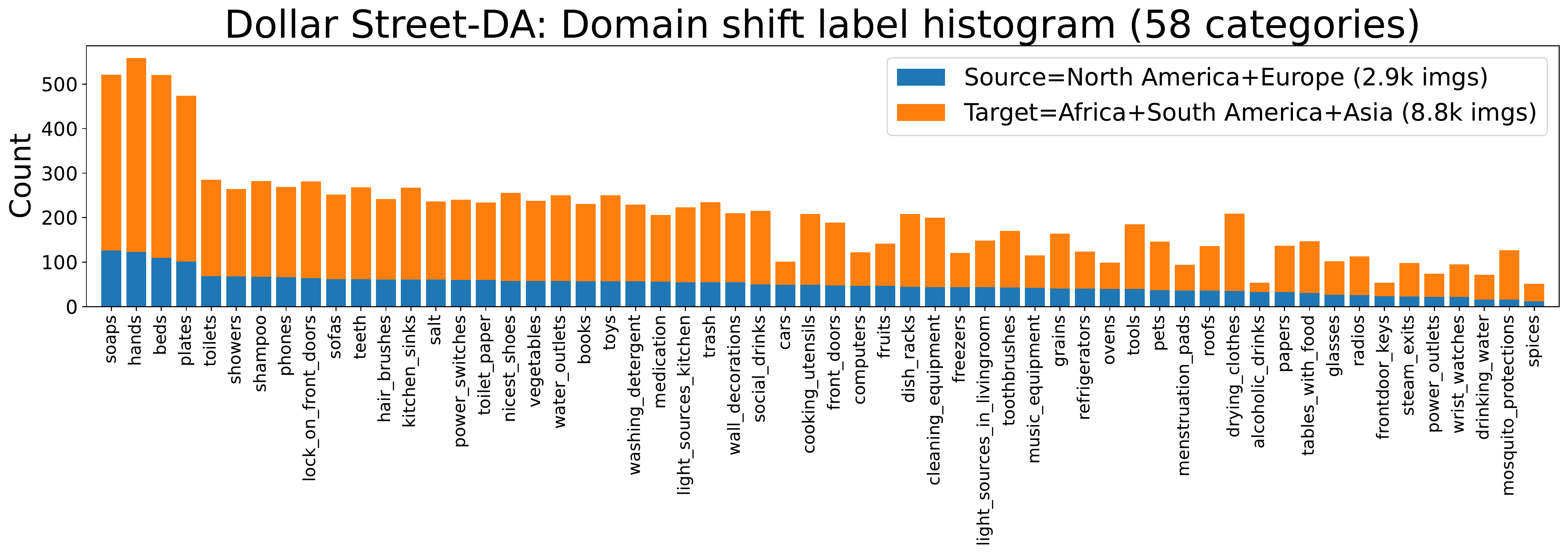}
          \caption{Dollar Street-DA: \{North America, Europe\} $\veryshortarrow$ \{Asia, Africa, South America\} }    
          \label{fig:dstreet}
          \end{subfigure}
          \begin{subfigure}[t]{0.9\textwidth}
            \vskip 0pt 
            \includegraphics[width=\textwidth]{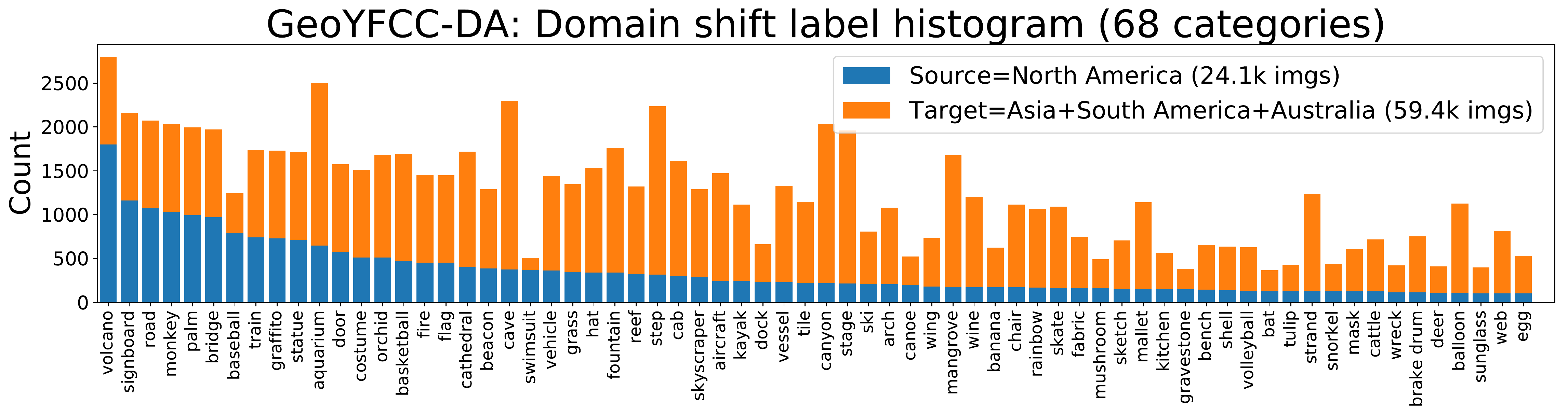}
            \vspace{10pt}
            \caption{GeoYFCC-DA: \{North America\} $\veryshortarrow$ \{Asia,  Australia, South America\} }
            \label{fig:geoda}
            \end{subfigure}
        \caption{Source (blue) and target (orange) label histograms for proposed Geographic DA shifts based on \textbf{(a)} Dollar Street~\cite{dstreet} \textbf{(b)}: GeoYFCC~\cite{dubey2021adaptive}.}
        \label{fig:histograms}
      \end{figure*}
     
    \begin{enumerate}
        \item To study the Geographical DA problem, we propose two adaptation shifts curated from the Dollar Street~\cite{dstreet} and GeoYFCC~\cite{dubey2021adaptive} datasets. 
        \item We validate the existence of context and subpopulation shift within these shifts, and experimentally verify that they pose significant challenges to model transfer.
        \item We benchmark the performance of representative domain adaptation techniques from the literature on these shifts and find them to achieve limited success, illustrating the need for specialized adaptation solutions for Geographical DA.
    \end{enumerate}

\section{Related Work}
\label{sec:relwork}

\noindent\textbf{Unsupervised domain adaptation (UDA).} UDA seeks to transfer a model trained on a labeled source to an unlabeled target domain, primarily via minimizing domain discrepancy statistics~\cite{long2015learning}, domain-adversarial learning~\cite{ganin2015unsupervised}, or self-training~\cite{prabhu2021sentry}. We formulate the problem of transferring trained image classification models to images from unseen geographies as UDA, and study how standard DA methods fare in this setting.

\begin{figure*}[t]
    \centering
    \includegraphics[width=0.7\linewidth]{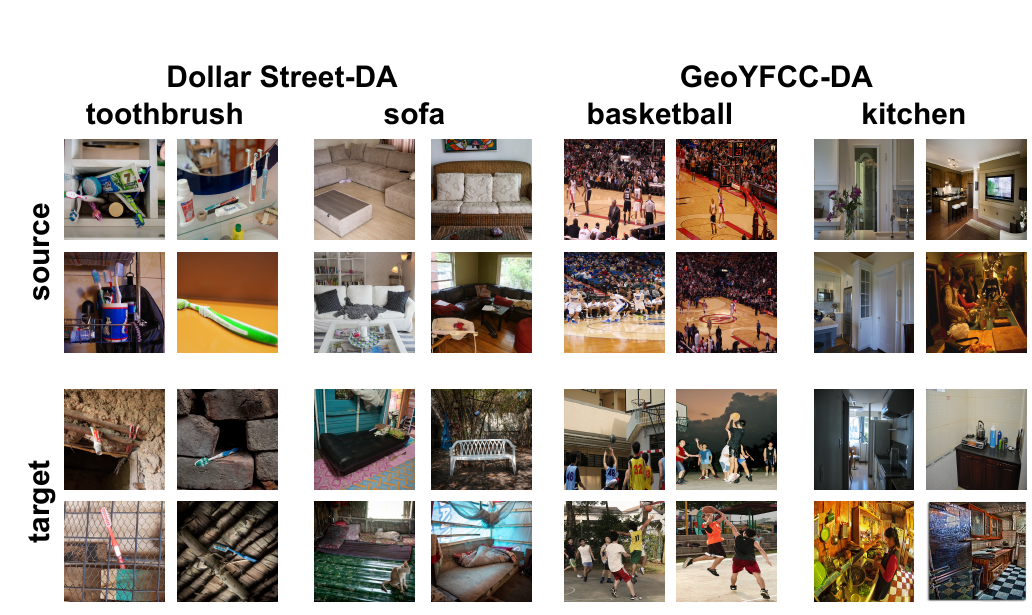}     
    \vspace{-10pt}   
    \caption{Context shift for select categories across domains. \textbf{Left:} Dollar Street-DA \textbf{Right:} GeoYFCC-DA.}
    \label{fig:context}
  \end{figure*}

\noindent\textbf{Geographical transfer learning.} Geographical transfer learning has received limited attention. Wang~\emph{et al.}~\cite{wang2020train} focus on cross-country adaptation of 3D object detectors, and propose a simple correction solution based on differences in average car sizes. Dubey~\emph{et al.} study the problem of geographical domain \emph{generalization}, and propose an adaptive solution that uses auxiliary target domain embeddings but unlike our setting does not allow training on unlabeled target data. Concurrent work~\cite{sagawa2021extending} extends the recently proposed WILDS~\cite{koh2021wilds} domain generalization benchmark to the unsupervised DA setting, and include one shift (FMoW~\cite{christie2018functional}) for geographical adaptation of models trained for land use prediction from satellite imagery. In contrast, we study geographical adaptation of object recognition models trained on standard internet imagery from the Dollar Street and YFCC datasets, which poses unique challenges of context and subpopulation shift.

\noindent\textbf{Context and subpopulation shift.} 
Singh~\emph{et al.}~\cite{singh2020don} study the problem of contextual biases learned by deep models based on frequently co-occuring categories. Some recent works~\cite{selvaraju2021casting,mo2021object} study the problem of minimizing contextual biases when learning self-supervised representations from scene-level imagery. Recent work proposes the BREEDS benchmark~\cite{santurkar2020breeds} to study model robustness against subpopulation shift---the ability to generalize to novel data subclasses not seen during training. Cai~\emph{et al.}~\cite{cai2021theory} study propose a input-consistency based label propagation algorithm to overcome subpopulation shift. To our knowledge, we are the first to study these challenges in the context of geographical DA.

\newcommand{\imgInputSpace}{\mathcal{X}}
\newcommand{\imgInput}{\mathbf{x}}
\newcommand{\imgInputTgt}{\imgInput_{\target}}
\newcommand{\imgInputTgtRescale}{\tilde{\imgInput}_{\target}}
\newcommand{\outputSpace}{\mathcal{Y}}
\newcommand{\outputLbl}{y}
\newcommand{\pseudoLbl}{\hat{\outputLbl}}
\newcommand{\numCls}{C}
\newcommand{\outputProb}{\mathbf{p}}

\newcommand{\Va}{V}
\newcommand{\Vb}{\tilde{V}}

\section{Benchmarks and Challenges}
\label{sec:approach}

We first present our two shifts for geographical domain adaptation curated from the Dollar Street and GeoYFCC datasets. We describe our curation process and analyze the characteristics of each geographical domain shift. We then describe and visualize the context and subpopulation shift present in these benchmarks. 

\subsection{Benchmarks}
\label{subsec:benchmarks}

\noindent\textbf{Dollar Street-DA.} The Dollar Street dataset was collected as part of the GapMinder project with the aim of using ``photos as data to kill country stereotypes''. It contains photographs and videos of everyday objects from peoples' homes spanning 66 unique countries. We restrict our study to image data and download images belonging to 128 unique categories. We filter out categories that  are scene-level or too broad (``agriculture lands'', ``play areas'') or abstract / subjective (``most loved items'', ``things I dream of having''), as well as categories with less than 50 images, resulting in 62 categories. We further deduplicate the dataset and merge some highly similar categories (e.g., ``plates of food'' and ``plates''), leaving us with images of 58 unique and distinct curated categories from 60 countries. We set up an adaptation problem from North America and Europe as source (2930 images) and Africa, South America, and Asia as target (8813 images). Fig.~\ref{fig:dstreet} presents a label histogram of each domain. 

\noindent\textbf{GeoYFCC-DA.} The GeoYFCC dataset~\cite{dubey2021adaptive} contains 1.1 million images from 62 countries curated from the subset of YFCC100M~\cite{thomee2016yfcc100m} images with geotags that were then automatically labeled based on keyword matching of image tags against ImageNet-5K categories excluding those in ILSVRC12~\cite{russakovsky2015imagenet}. We create an adaptation problem from countries in North America as source and countries in Asia, South America, and Australia as our target domain. Due to the automatic labeling pipeline, we notice a large amount of label noise in the dataset and take two measures to curate the dataset further: i) We train a ResNet50~\cite{he2016deep} model on the source domain and measure heldout test accuracy, and only retain classes with $>25\%$ accuracy. ii) We manually inspect 100 random qualitative examples from the source and target domains for the remaining categories and exclude categories with significant label noise. At the end of this process, we select 68 categories with 24.1k images in the source domain and 59.4k images in the target domain. See Fig.~\ref{fig:geoda} for a label histogram of each domain.

\begin{figure*}[t]
  \centering
  \includegraphics[width=0.6\linewidth]{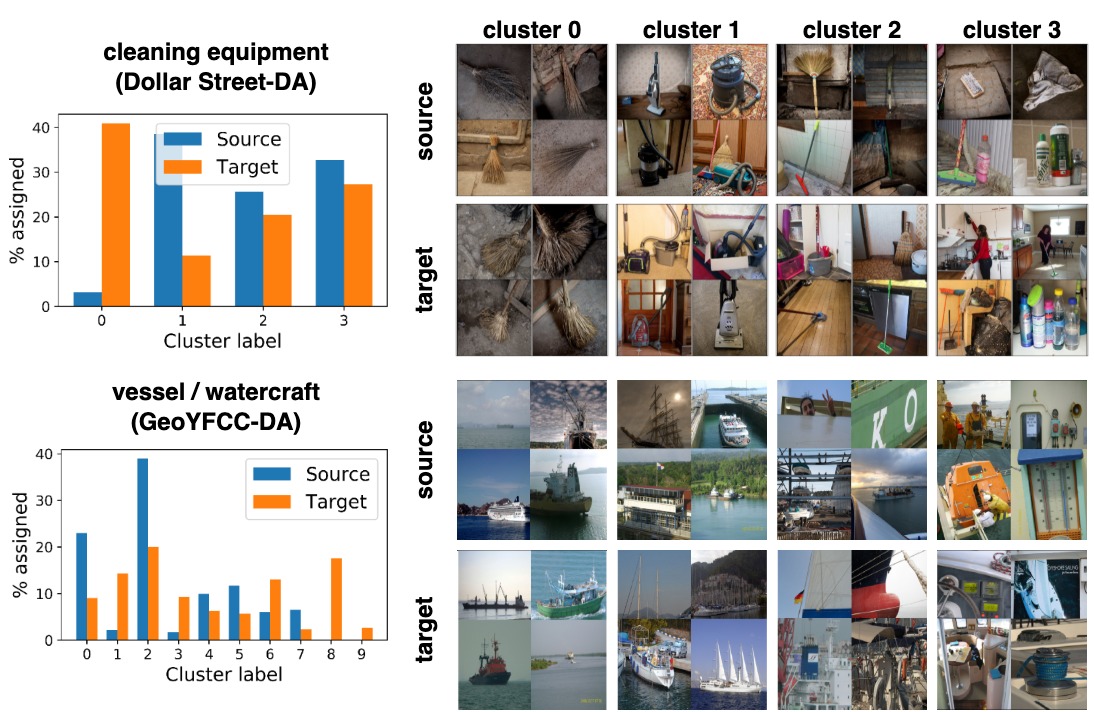}        
  \caption{Subpopulation shift for select categories across domains. We plot normalized cluster assignments per-domain as approximate subpopulation distributions -- blue denotes within-category subpopulation distribution on source and orange denotes target. As seen, subpopulation distributions shift significantly across domains. On the right we visualize random images from some of the discovered clusters, and verify that they generally correspond to distinct subpopulations.}
  \label{fig:subpop}
\end{figure*}

\subsection{Challenges: Context and subpopulation shift}
\label{subsec:challenges}

\noindent\textbf{Notation.} \noindent Let $\mathcal{X}$ and $\mathcal{Y}$ denote input and ouput spaces, with the goal being to learn a convolutional neural network $\model: \mathcal{X} \to \mathcal{Y}$ parameterized by $\modelparams$. In unsupervised DA we are given access to labeled source examples $(\mathbf{x}_\source, y_\source) \sim \mathcal{P}_\source(\mathcal{X}, \mathcal{Y})$, and unlabeled target examples $\mathbf{x}_\target \sim \mathcal{P}_\target(\mathcal{X})$, where $\source$ and $\target$ denote the source and target domains. The goal is to maximize model accuracy on the target domain, and we consider adaptation of models trained to perform $K$-way object recognition: the inputs $\mathbf{x}$ are images, and labels $y$ are categorical variables $y \in \{1, 2, .. , \numclasses \}$.

\noindent\textbf{Data and label distribution shift.} As in conventional domain adaptation, geographical DA also presents data distribution shift ($\mathcal{P}_\source(\mathbf{x}) \neq \mathcal{P}_\target(\mathbf{x})$) as object appearances change across geographies (see Fig.~\ref{fig:teaser}), and label distribution shift ($\mathcal{P}_\source(y) \neq \mathcal{P}_\target(y)$), as task label distributions change across domains (see Fig.~\ref{fig:histograms}).

In addition, geographical adaptation presents two new challenges: context and subpopulation shift.

\noindent\textbf{Context shift.} We define context $c(\mathbf{x})$ for image $\mathbf{x}$ with label $y$ as the \emph{task-irrelevant} information in the image---this loosely corresponds to the background or surroundings of the object of interest. We define context shift as $\mathcal{P}_\source(c(\mathbf{x})|y) \neq \mathcal{P}_\target(c(\mathbf{x})|y) $, representing a change in object context across geographical domains. 

In Fig.~\ref{fig:context} we show qualitative examples of context shift within our proposed Dollar Street-DA and GeoYFCC-DA shifts for a few categories. For example, we find that in Dollar Street-DA, most ``toothbrush'' images in the source domain tend to be photographed inside bathrooms, whereas the surroundings in the target domains are \emph{significantly} more diverse (\emph{e.g.} walls and roofs). We see similar trends in the GeoYFCC-DA shift (\emph{e.g.} indoor v/s outdoor basketball games). As deep neural networks are known to often employ ``shortcut learning''~\cite{geirhos2020shortcut} of potentially spurious features (\emph{e.g.} object backgrounds) to make predictions, we hypothesize (and experimentally verify in Sec.~\ref{subsec:transfer}) that such a context shift will present a challenge to visual recognition models deployed in new geographies.

\noindent\textbf{Subpopulation shift.} We define subpopulation shift as $\mathcal{P}_\source(\mathbf{x}|y) \neq \mathcal{P}_\target(\mathbf{x}|y)$, representing a change in within-category distribution across domains.

In Fig.~\ref{fig:subpop}, we show examples of subpopulation shift in the Dollar Street-DA and GeoYFCC-DA benchmarks. In the absence of subpopulation-level annotations, we use a simple strategy to obtain \emph{approximate} annotations: we use a pretrained model (ResNet50~\cite{he2016deep} trained on ImageNet) to extract features for source and target images of a given category, and perform agglomerative clustering on the combined set of embeddings. We then use the inferred cluster assignments as subpopulation annotations. We also plot the normalized within-class distribution of cluster assignments on the source and target domains, and measure the Wasserstein distance between the two as a measure of the degree of subpopulation shift.

As seen, this simple strategy discovers distinct clusters corresponding to semantically distinct subpopulations: \emph{e.g.} for ``cleaning equipment'' on Dollar Street-DA we discover separate clusters roughly corresponding to brooms, vaccuum cleaners, mops, and miscellaneous cleaning items. Crucially, we find that the \emph{intra-class distribution} of many categories changes significantly across geographies (\emph{e.g.} brooms make up a significantly larger proportion of cleaning equipment in the target domain than in the source).

\begin{wrapfigure}{r}{0.4\linewidth}
  \vspace{-15pt}
  \includegraphics[width=\linewidth]{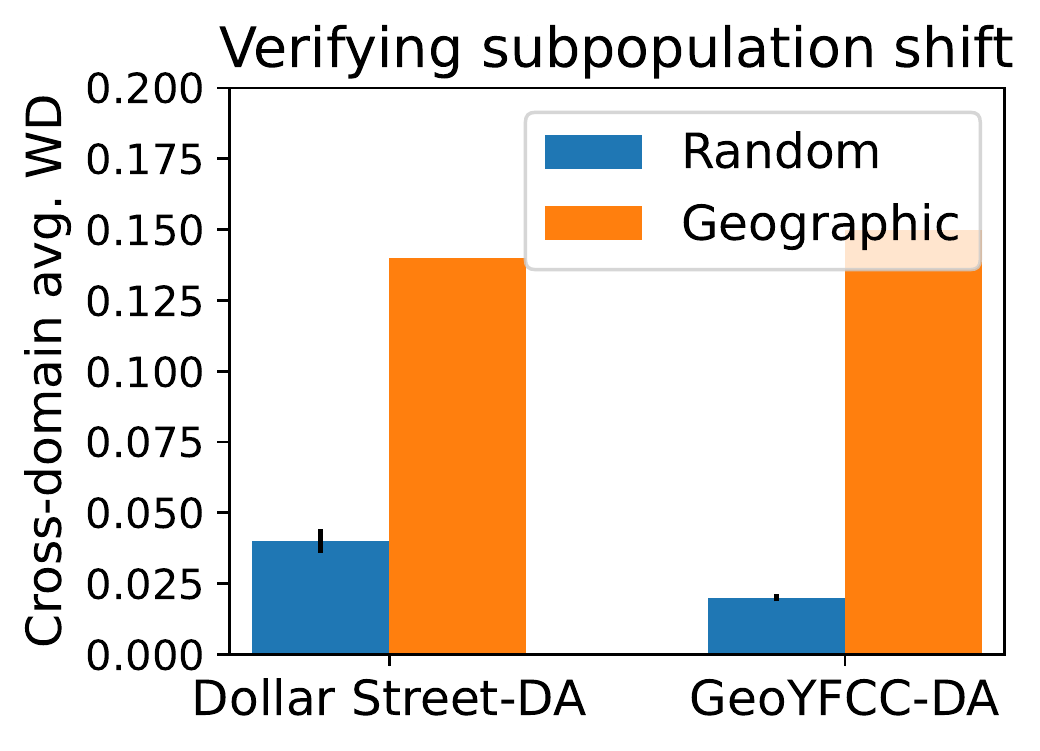}        
  \vspace{-15pt}
  \caption{Verifying subpopulation shift.}
  \label{fig:subpop_wd}
\end{wrapfigure}
To quantitatively validate the subpopulation shift, we compute the \emph{degree} of per-class subpopulation shift: for each class, we compute the normalized subpopulation distribution per-domain (as visualized in Fig.~\ref{fig:subpop}, left), measure the cross-domain Wasserstein distance, and average across classes. 
Fig.~\ref{fig:subpop_wd} presents results for this measure for our proposed geographical shifts versus randomly constructed source and target domains of the same size, for the Dollar Street and GeoYFCC datasets. As seen, geographical shifts lead to \emph{significantly} higher subpopulation shift.

\section{Experiments}
\label{sec:experiments}

\subsection{Setup}
\label{subsec:setup}

To account for label imbalance, we report per-class average accuracy as our metric. As described in Sec.~\ref{sec:approach}, we consider adaptation on two shifts:

\noindent\textbf{Dollar Street-DA}: We consider images from North America and Europe as the source domain (2.9k) and images from Asia, Africa, and South America as the target domain (8.8k). 

\noindent\textbf{GeoYFCC-DA}: We consider images from North America as the source domain (24.1k) and images from Asia, Australia, and South America as the target domain (59.4k). 

On both shifts we create a 90\%-10\% train-test split on the source domain and report transfer performance on 20\% heldout target data, but use the entire target dataset for unsupervised adaptation. On Dollar Street-DA, due to the relatively small size of the dataset, we report performance mean and  standard deviation over three experimental runs.

\begin{table}[t]
  \begin{center} 
  \resizebox{\columnwidth}{!}{
  \begin{tabular}{l c c}
  \toprule
  {\bf Method} & {\bf Dollar Street-DA} & {\bf GeoYFCC-DA} \\
  \midrule    
source & {54.66\scriptsize $\pm$0.62} & 42.88 \\
target oracle* & {67.73\scriptsize $\pm$ 0.30} & 56.78 \\
\midrule
 MMD~\cite{long2015learning} & {55.77\scriptsize $\pm$0.75} & 43.53 \\
 DANN~\cite{ganin2015unsupervised}  & {54.80\scriptsize $\pm$0.38} & 42.64 \\
 SENTRY~\cite{prabhu2021sentry}  & 55.73\scriptsize $\pm$0.34 & 42.58 \\
 SST  & {58.71\scriptsize$\pm$0.53} & 45.22 \\
  \bottomrule
  \end{tabular}}
  \vspace{-10pt}
  \caption{Average accuracy on the target test set (20\%) for the Dollar Street-DA and GeoYFCC-DA shifts. * denotes that the target oracle was trained on target data non-overlapping with the test set (80\%) whereas DA methods were adapted without labels on the entire target dataset.}
  \label{tab:da_baselines}
  \end{center}
\end{table}

\subsection{Domain Adaptation Baselines} 
\label{subsec:da}

We benchmark 4 representative DA methods from the literature on our Geographical DA shifts: one domain discrepancy-based method, one domain adversarial method, and two self-training based methods.

\noindent\textbf{1) MMD~\cite{long2015learning}:} Aligns domains by computing mean source and target embeddings in a reproducing kernel hilbert space and minimizing their distance as a maximum mean discrepancy measure.

\noindent\textbf{2) DANN~\cite{ganin2015unsupervised}:} Domain-adversarial neural networks adversarially learn a domain discriminator to distinguish source and target features against a feature encoder that is trained to fool the discriminator.

\noindent\textbf{3) SENTRY~\cite{prabhu2021sentry}:} SENTRY measures model predictive consistency across randomly augmented versions of each target image and selectively increases predictive entropy on highly consistent instances, while decreasing it on highly inconsistent ones. SENTRY also uses pseudolabel-based approximating class balancing on the target domain, and employs a slightly modified ResNet50 architecture~\cite{chen2018closer}.

\noindent\textbf{4) Selective Self-training (SST):} We implement a simplified self-training baseline that self-trains against predicted labels on target instances on which the model is atleast 90\% confident, and also employs pseudolabel-based target class balancing and a modified ResNet50 architecture~\cite{chen2018closer}. 

Finally, we also report performance for a \emph{target oracle} that is trained in a supervised fashion on the target domain. The target oracle is meant to represent a performance \emph{upper bound} in the absence of domain shift.

\noindent\textbf{Implementation details.} We use a ResNet50~\cite{he2016deep} as our CNN architecture, and for SENTRY and SST use a modified few-shot variant~\cite{chen2018closer} that replaces the last linear layer with a K-way (where K is the number of classes) fully-connected layer without bias. Input activations to this layer are $L_2$-normalized and its output is passed into a softmax layer with a temperature of 0.05. For optimization, we use Adam~\cite{kingma2014adam} with a learning rate of 0.001 and weight decay of $5e^{-4}$. We use a batch size of 128 across experiments. All DA methods are applied to a source model that is first trained using supervised training on the source domain for 50 epochs. We employ 100 epochs of adaptation on Dollar Street-DA and 40 epochs on GeoYFCC-DA. All methods additionally minimize a supervised cross-entropy loss on source labels during adaptation (with a loss weight of 1.0 and 0.1 on Dollar Street-DA and GeoYFCC-DA respectively). To combat label distribution shift, we follow Tan~\emph{et al.}~\cite{tan2020class} and use class-balanced sampling on the source domain across experiments.

\begin{figure*}[t]
  \centering
  \begin{subfigure}[t]{0.9\textwidth}
      \vskip 0pt 
    \includegraphics[width=\textwidth]{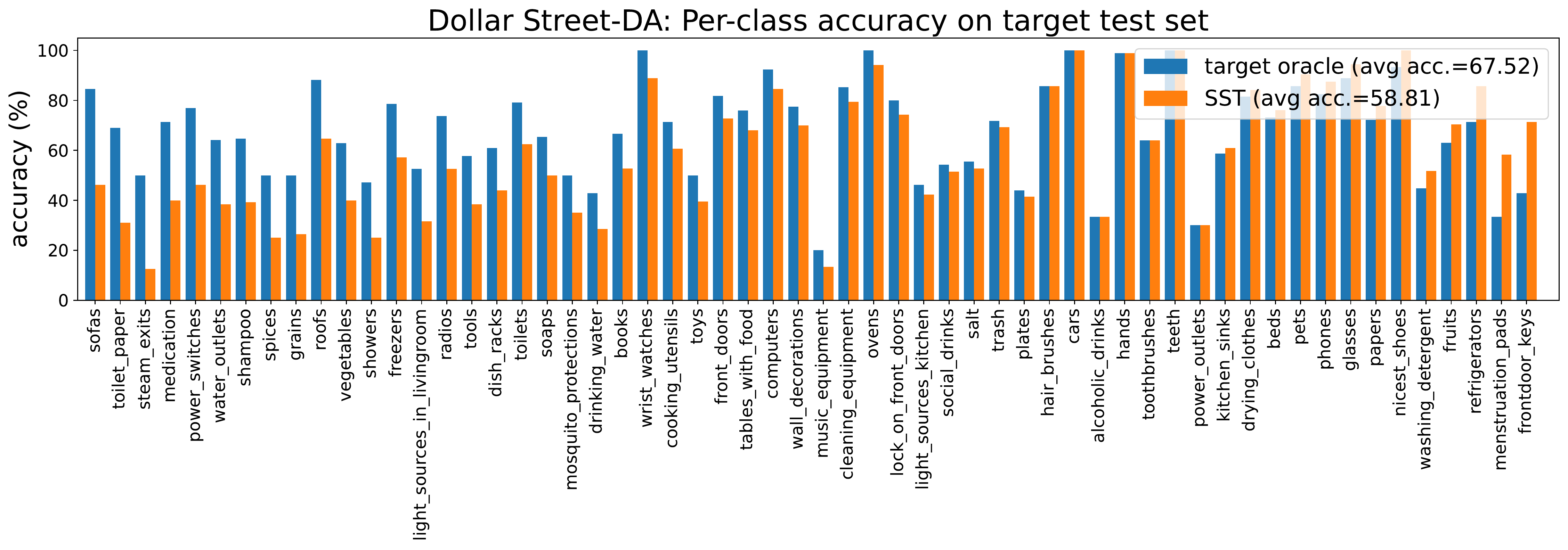}
    \caption{Dollar Street-DA}    
    \label{fig:dstreet_transfer}
    \end{subfigure}
    \begin{subfigure}[t]{0.9\textwidth}
      \vskip 0pt 
      \includegraphics[width=\textwidth]{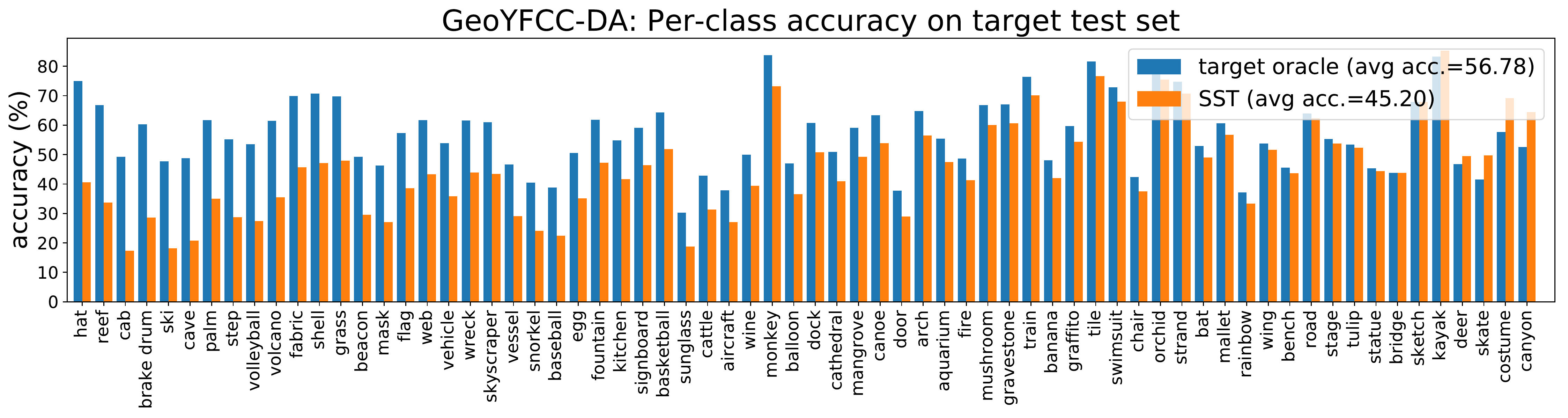}
      \vspace{10pt}
      \caption{GeoYFCC-DA}
      \label{fig:geoda_transfer}
      \end{subfigure}
      \vspace{-10pt}
  \caption{Per-category target test accuracy for a model trained on \textbf{blue:} target train set and \textbf{orange:} source train set and adapted to the target domain via the SST method. Categories are ordered in decreasing order of accuracy drop.}
  \label{fig:transfer}
\end{figure*}

\subsection{Results} 
\label{subsec:results}

Table~\ref{tab:da_baselines} presents average accuracy on the target test set for the Dollar Street-DA and GeoYFCC-DA shifts. We observe:

\noindent$\triangleright$ \textbf{Geographical shifts lead to significant performance drops (Row 1 v/s 2)}. As seen, the target oracle achieves 67.7\% and 56.78\% whereas the source model achieves a significantly lower performance of 54.66\% (-13.1\%) and 42.88\% (-13.9\%). Clearly, geographical variations pose a significant challenge to transfer. 

\noindent$\triangleright$ \textbf{DA methods offer limited improvements (Rows 3-6)}. All the benchmarked methods achieve limited success at geographical adaptation, sometimes performing no better than the source model. Surprisingly, we find the simple SST method to achieve the best (albeit small) improvement over the source model (+4.1\% / +2.3\%), but still well short of the target oracle (-9\% / -11.6\%). Altogether, these results indicate the need for specialized adaptation solutions for the Geographical DA problem.
 
\subsection{Analyzing failure modes} 
\label{subsec:transfer}

We now analyze the performance of our best-performing SST model and contrast it with the target oracle.

\begin{figure*}[t]
  \centering
    \includegraphics[width=\textwidth]{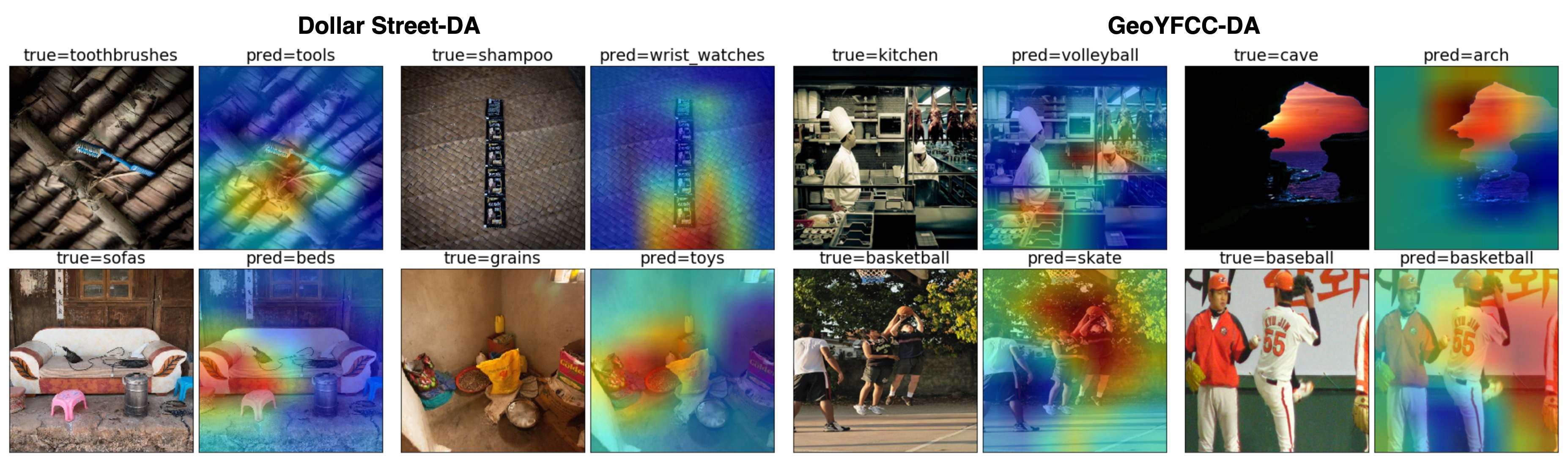}    
    \vspace{-10pt}
  \caption{Context shift as a failure mode for Geographical DA. We visualize incorrect predictions from the best-performing SST model on the target test set alongside visual explanations generated with GradCAM~\cite{selvaraju2017grad} for Dollar Street-DA (left) and GeoYFCC-DA (right). As seen, the model frequently attends to spurious background features and makes incorrect predictions.}
  \label{fig:context_fm}
\end{figure*}

\noindent\textbf{Per-class performance.} In Fig.~\ref{fig:transfer} we show per-category accuracies on the target test set for i) \textbf{blue:} a \emph{ target oracle} model trained on the target train set and ii) \textbf{orange:} the best performing SST model which was trained on the source train set and adapted to the entire target domain. We order categories in descending order of accuracy drop. As seen, the SST model performance lags behind the target oracle on most categories. We further analyze failure modes arising from context shift and subpopulation shift:

\noindent \textbf{Context shift.} In Fig.~\ref{fig:context_fm} we show examples of model errors on the target domain arising from context shift. We also visualize GradCAM~\cite{selvaraju2017grad} explanations along side each image, and find that, in most cases, the model makes erroneous predictions on target images with contexts that are uncommon in the source domain (see Fig.~\ref{fig:context} for source examples) while fixating on spurious background features.  

\noindent \textbf{Subpopulation shift.} To verify that subpopulation shift is a failure mode, we measure per-class average subpopulation accuracy, using the approximate subpopulation-level annotations described in Sec.~\ref{subsec:challenges}. We then measure the correlation between the drop in per-class average subpopulation accuracy for the SST model compared to the target oracle (that does not experience subpopulation shift), against the degree of subpopulation shift measured via the per-category average of cross-domain Wasserstein distance between subpopulation distributions. On the 40 classes with the highest subpopulation shift in Dollar Street-DA, we observe a Pearson correlation coefficient of 0.44, and 0.39 on GeoYFCC-DA.
This indicates the tendency of the adapted model to underperform to a larger degree---when compared to the target oracle---on categories with high subpopulation shift.

\vspace{-10pt}
\section{Limitations \& Conclusion}
\label{sec:conclusion}

Our work has some important limitations: we do not consider semantic drift, where the meaning of a category itself may change across geographies \emph{e.g.} a ``chair'' in one country might be considered as a ``sofa'' in another. We also restrict our study to adaptation across continent-level shifts, but analyzing shifts across a more fine-grained (\emph{e.g.} country) level is also potentially valuable. Moreover, variation in visual appearance within a geography can sometimes be larger than that across geographies, due to other confounding factors like income and demographics---we do not study these differences here. Finally, in the absence of human annotations, we are restricted to using inferred subpopulation annotations for our analysis.

To summarize, we studied the problem of domain adaptation of trained object recognition models to new geographies, and investigated the effectiveness of off-the-shelf adaptation methods. We proposed two shifts to study this problem and demonstrated the existence of two unique challenges: cross-domain context shift and subpopulation-shift. We found existing DA methods to offer limited success at Geographical DA, suggesting the need for future work to develop specialized adaptation solutions for this important but understudied problem.

{\small
\bibliographystyle{ieeetr}
\bibliography{main}
}
\newpage

\end{document}